%% file: _main.tex
\documentclass[10pt]{article} 

\usepackage[accepted]{rlj} 

\usepackage{amssymb}            
\usepackage{mathtools}          
\usepackage{mathrsfs}           
\usepackage{graphicx}           
\usepackage{subcaption}         
\usepackage[space]{grffile}     
\usepackage{url}                
\usepackage{lipsum}             
\usepackage{cleveref}

\newcommand{\stateSpace}{\mathcal{X}}
\newcommand{\actionSpace}{\mathcal{A}}
\newcommand{\transFn}{\mathcal{P}}
\newcommand{\rewFn}{\mathcal{R}}
\newcommand{\reals}{\mathbb{R}}
\newcommand{\dataset}{\mathcal{D}}

\title{Optimistic critics can empower small actors}

\setrunningtitle{Optimistic critics can empower small actors}

\author{Olya Mastikhina\textsuperscript{ $\dagger$, 1,2}, Dhruv Sreenivas\textsuperscript{$\dagger$, 1,2}, Pablo Samuel Castro\textsuperscript{1,2,3}}

\emails{\{olya.mastikhina,dhruv.sreenivas,pablo-samuel.castro\}@mila.quebec}

\affiliations{
$^{1}$\textbf{Mila - Qu\'ebec AI Institute}\\
$^{2}$\textbf{Universit\'e de Montr\'eal}\\
$^{3}$\textbf{Google DeepMind}
\par 
$^\dagger$ Equal contribution.
}

\input{1_contributions.tex}

\begin{document}

\maketitle  

\input{0_abstract_or_summary}
\input{2_introduction}
\input{3_related_work_background}

\input{4_results}

\input{4.5_fixes}

\input{5_conclusion}

\bibliography{refs}
\bibliographystyle{rlj}

\input{6_appendix}

\end{document}

%% file: 1_contributions.tex
\contribution{
    We show that reducing the size of the actor in actor-critic methods can lead to degraded performance and increased overfitting in the critic.
}
{
    Prior work suggests that actors require less capacity than critics in actor-critic algorithms \citep{mysore_honey_2021}, and that asymmetric training with smaller actors can be beneficial for real-world applications \citep{degrave22magnetic}.
}
\contribution{
    We demonstrate that performance degradation and critic overfitting is largely due to poorer data collection, and this arises due to value underestimation.
}
{
    This is somewhat surprising, as it stands in contrast to the {\em over-estimation} that's commonly addressed in many popular algorithms \citep{hasselt2010double,hasselt2016deep,fujimoto2018addressing}. However, other papers have shown that underestimation can be an issue with the actor-critic algorithms that address overestimation \citep{NEURIPS2019_a34bacf8, 10.3389/fnbot.2022.1081242, He_2020}.
}

\contribution{
    We explore a number of approaches for mitigating the value underestimation and find the most effective one to be replacing the $\min$ term with an average or $\max$ term when combining the value estimates of two critics (as done in SAC).
}
{
    Taking the minimum of two estimated $Q$-values will, by definition, be conservative; indeed, the idea was originally proposed to deal with over-estimation \citep{hasselt2010double}. Prior work has shown that resetting or regularizing the critic in particular improves plasticity \citep{ma_revisiting_2023, nikishin_primacy_2022, liu2021regularization} and can help mitigate value-estimation issues, particularly in the case of layer normalization \citep{nauman_overestimation_2024}.
}

\keywords{Deep reinforcement learning, actor-critic, asymmetric actor-critics, exploration, value underestimation, data collection} 

\summary{Actor-critic methods have been central to many of the recent advances in deep reinforcement learning. The most common approach is to use {\em symmetric} architectures, whereby both actor and critic have the same network topology and number of parameters. However, recent works have argued for the advantages of {\em asymmetric} setups, specifically with the use of smaller actors. We perform broad empirical investigations and analyses to better understand the implications of this.}

%% file: 0_abstract_or_summary.tex
\begin{abstract}
Actor-critic methods have been central to many of the recent advances in deep reinforcement learning. The most common approach is to use {\em symmetric} architectures, whereby both actor and critic have the same network topology and number of parameters. However, recent works have argued for the advantages of {\em asymmetric} setups, specifically with the use of smaller actors. We perform broad empirical investigations and analyses to better understand the implications of this and find that, in general, smaller actors result in performance degradation and overfit critics. Our analyses suggest {\em poor data collection}, due to value underestimation, as one of the main causes for this behavior, and further highlight the crucial role the critic can play in alleviating this pathology. We explore techniques to mitigate the observed value underestimation, which enables further research in asymmetric actor-critic methods.

\end{abstract}

%% file: 2_introduction.tex
\section{Introduction}

Actor-critic (AC) algorithms are a fundamental part of deep reinforcement learning (RL), with various AC methods achieving state-of-the-art performance in complex discrete control \citep{espeholt_impala_2018} and continuous control \citep{haarnoja_soft_2018} tasks. In these approaches, the actor interacts with the environment to collect data and to optimize a mapping of states to actions with the guidance of the critic, while the critic learns a value function with the collected data to guide the actor's learning. These symbiotic, but differing, roles have been traditionally implemented with either coupled or matching ("symmetric") neural network architectures \citep{haarnoja2018sac2, yarats2021image}; however, recent work suggests that the actor requires less capacity and can be significantly reduced relative to the critic \citep{mysore_honey_2021}.

As only the actor is used during inference, reducing the size of the actor while keeping a bigger critic offers several advantages for real-world applications. A smaller actor reduces inference costs, which is beneficial for resource-constrained applications such as robotics, where fast computations are essential for real-time performance \citep{hu_toward_2024, schmied_large_2025}, and inference time is a bottleneck for deployment \citep{firoozi_foundation_2024}. 
Decoupling the size of the actor from the critic allows for bigger critics that can fully leverage data available in simulators for learning complex tasks without then affecting inference costs. This approach has recently been successfully applied to training an RL agent for the magnetic control of tokamak plasmas for nuclear fusion - an application that requires particularly fast computation speeds \citep{degrave22magnetic}.

Beyond computational constraints, another barrier to real-world deployment is interpretability and the incorporation of safety constraints, which are particularly important for safety-critical applications like autonomous driving \citep{tang_deep_2024, xu_benchmarking_2023, xiao_motion_2022}. 
Smaller actors tend to generate simpler policies which are easier to interpret \citep{fan_interpretability_2021, li_interpretable_2022}. While distillation is another promising approach for generating compact policies for real-world deployment \citep{Hinton2015DistillingTK, rusu2016policydistillation, liu2024visual}, direct training makes the incorporation of safety and functional constraints simpler and more reliable.

Despite their apparent advantages, there has been little work in developing an understanding of how to properly train asymmetric AC methods with smaller actors, as well as how the actor-critic relationship is affected by this asymmetry. In this paper, we address this gap by performing a broad empirical investigation with the Soft Actor-Critic \citep[{SAC};][]{haarnoja2018sac2} and Data-Regularized Q \citep[{DrQ};][]{yarats2021image} agents in the physics-simulated DeepMind Control suite \citep[DMC;][]{tassa_deepmind_2018, tunyasuvunakool_dm_control_2020} environments. We reduce the number of parameters in the actor (sometimes as far down as $1\%$ of its original size) and observe increased overfitting in the critics as actor size decreases. 
However, rather than this being a hard limitation due to capacity loss, our analyses suggest that this performance drop can mostly be attributed to poorer data collection by the actor, which may be caused by pessimistic under-exploration problems with algorithms like SAC and DrQ that compute the minimum of Q value estimates \citep{NEURIPS2019_a34bacf8, haarnoja2018sac2, yarats2021image}. Notably, we find that simply alleviating value under-estimation in the critics can drastically improve performance. We show a similar mitigation effect for a drop in performance caused by the actor receiving limited information, suggesting assisting constrained actors with optimism may be a general strategy for conservative AC methods. 

The paper is organized as follows: in \cref{sec:preliminaries}, we lay the groundwork and explain our experimental setup. In \cref{sec:impactOfSmallActors}, we show the performance effects when naively reducing a smaller actor across a variety of state-based and image-based continuous control tasks, and analyze what could be the cause of performance differences. In \cref{sec:empoweringSmallActors}, we focus on interventions that can gain back performance, specifically focusing on bias correction and value function underestimation. Finally, we conclude with discussions and avenues for future work in \cref{sec:conclusion}.

%% file: 3_related_work_background.tex
\section{Preliminaries}
\label{sec:preliminaries}

Reinforcement learning (RL) agents learn by interacting with an environment, which is typically formulated as a Markov decision process (MDP) $\langle \stateSpace, \actionSpace, \transFn, \rewFn \rangle$ \citep{puterman1994mdps}. Here, $\stateSpace$ denotes the agent state space; $\actionSpace$ is the set of actions available to the agent; $\transFn:\stateSpace\times\actionSpace\rightarrow\Delta(\stateSpace)$ are the transition dynamics with $\transFn(x^{\prime} \mid x, a)$ indicating the probability of transitioning to state $x^{\prime}\in\stateSpace$ after selecting action $a\in\actionSpace$ from state $x\in\stateSpace$; $\rewFn:\stateSpace\times\actionSpace\rightarrow\reals$ is the reward function, where $\rewFn(x, a)$ denotes the reward received after performing action $a$ from state $x$. An agent's behavior is quantified by a \emph{policy} $\pi:\stateSpace\rightarrow\Delta(\actionSpace)$, where $\pi(a \mid x)$ denotes the probability of selecting action $a$ when in state $x$. The estimated returns of a policy $\pi$ from state $x$ are quantified via the (recursive) \emph{value function} $V^{\pi}(x) := \mathbb{E}_{a \sim \pi(\cdot \mid x)}\left[ \rewFn(x, a) + \gamma\mathbb{E}_{x'\sim\transFn(\cdot \mid x, a)}V^{\pi}(x')\right]$, where $\gamma\in [0,1)$ is a discount factor that discourages waiting too long before obtaining rewards. We can define the \emph{state-action value function} $Q^\pi(x, a)$, which quantifies the value of taking an arbitrary action $a$ from state $x$, and then following $\pi$ afterwards: $Q^{\pi}(x, a) := \rewFn(x, a) + \gamma \mathbb{E}_{x'\sim\transFn(\cdot \mid x, a)}V^{\pi}(x')$. One can easily see that $V^\pi(x) = \mathbb{E}_{a \sim \pi(\cdot \mid x)} Q^\pi(x, a)$. The goal of RL is to find an \emph{optimal} policy $\pi^*$ which maximizes returns, in the sense that $V^{\pi^*} \geq V^{\pi}$ for all $\pi$. There are a number of techniques for learning optimal policies, most of which alternate between \emph{policy evaluation} and \emph{policy improvement}. Policy evaluation seeks to estimate the value function of a policy $\pi$, which is primarily done by minimizing temporal-difference (TD) errors:
\begin{align}
    TD^\pi(x, a, x') = \vert Q^\pi(x, a) - (\rewFn(x, a) + \gamma V^\pi(x')) \vert \label{eqn:tdError}.
\end{align}

Policy improvement then follows by directly maximizing this Q function, either via a standard arg-max over actions or gradient ascent. Actor-critic methods operate by maintaining separate estimates of $\pi_{\theta}$ (the actor) and $Q_{\phi}$ (the critic), which are used in each of the learning objectives; in deep RL, these functions are approximated by neural networks, parameterized by $\theta$ and $\phi$, respectively. Given a dataset $\dataset$ of transitions (often stored in a replay buffer), Soft Actor-Critic \citep[{\bf SAC}; ][]{haarnoja_soft_2018,haarnoja2018sac2} optimizes the actor and critic by minimizing the following losses:
\begin{align}
    J_Q(\phi) & = \mathbb{E}_{x, a, x' \sim\dataset} \left[ \frac{1}{2}\left(Q_{\phi}(x, a) - \left( \rewFn(x, a) + \gamma V_{\bar{\phi}}(x') \right) \right)^2 \right] \label{eqn:sac_jq} \\
    J_{\pi}(\theta) & = \mathbb{E}_{x\sim\dataset}\left[ \mathbb{E}_{a\sim\pi_{\theta}(\cdot \mid x)}\left[ \alpha\log \pi_{\theta}(a \mid x) - Q_{\phi}(x, a) \right] \right] \label{eqn:sac_jpi}
\end{align}

In \cref{eqn:sac_jq}, $V_{\bar{\phi}}$ is the value function computed from $Q_{\phi}$ via $V_{\bar{\phi}}(x) = \mathbb{E}_{a\sim\pi_{\theta}(\cdot \mid x)}\left[ Q_{\bar{\phi}}(x, a) - \alpha \log\pi_{\theta}(a \mid x) \right] $, where $\bar{\phi}$ are delayed target parameters \citep{mnih2015human}, and $\alpha$ is a learned Lagrange multiplier (we exclude its parameterization for simplicity of exposition). In their practical implementation, \citet{haarnoja2018sac2} use two $Q$ value estimates with parameters $\phi_1$ and $\phi_2$, trained independently, and take their minimum in the update terms in equations \ref{eqn:sac_jq} and \ref{eqn:sac_jpi}, resulting in the following updated losses, with $V_{\bar{\phi}}(x) = \mathbb{E}_{a\sim\pi_{\theta}(\cdot \mid x)}\left[ \min_{i\in\{1,2\}}Q_{\bar{\phi_i}}(x, a) - \alpha \log\pi_{\theta}(a \mid x) \right]$:
\begin{align}
    J_Q(\phi_i) & = \mathbb{E}_{x, a, x' \sim\dataset} \left[ \frac{1}{2}\left(Q_{\phi_i}(x, a) - \left( \rewFn(x, a) + \gamma V_{\bar{\phi}}(x') \right) \right)^2 \right] \label{eqn:sac_jq_min} \\
    J_{\pi}(\theta) & = \mathbb{E}_{x\sim\dataset}\left[ \mathbb{E}_{a\sim\pi_{\theta}(\cdot \mid x)}\left[ \alpha\log \pi_{\theta}(a \mid x) - \min_{i\in\{1,2\}}Q_{\phi_i}(x, a) \right] \right] \label{eqn:sac_jpi_min}
\end{align}

It is important to note the interconnectedness of these losses: the actor influences the critic via the (soft) value function $V_{\bar{\phi}}$ used in \cref{eqn:sac_jq,eqn:sac_jq_min}, while the critic influences the actor via $Q_{\phi}$ in \cref{eqn:sac_jpi,eqn:sac_jpi_min}. Additionally, the actor influences the training dynamics of both given that it is in charge of data collection. Finally, note the use of the TD-error term in \cref{eqn:sac_jq,eqn:sac_jq_min}.

\subsection{Experimental setup}
We run our experiments on the DeepMind Control suite \citep[DMC;][]{tassa_deepmind_2018, tunyasuvunakool_dm_control_2020}, a suite of continuous control tasks that have been a staple of continuous-action reinforcement learning research. For any of the tasks, DMC can provide either low-dimensional features or pixel observations to the agents, while keeping the underlying transition and reward dynamics unchanged. Pixel-based observations are generally more challenging, as the MDP is partially observed \citep{pomdp, yarats2020improving}, but investigating both provides richer insights into the dynamics of the examined learning algorithms.

Due to computational limitations, the bulk of our analyses will be on feature-based tasks. For these, we use as baseline the default set up and parameters for DMC \citep{haarnoja2018sac2}. This consists of one actor network, two critic networks, and two critic target networks. The critic and target networks consist of two hidden layers of size 256 and output a one dimensional Q value estimate. By default, the actor consists of hidden layers of size 256, with two output layers that parameterize the mean and standard deviation of a Gaussian distribution squashed by a tanh function. The critic and actor networks are decoupled, in the sense that they share no parameters.

For pixel observations we use {\bf DrQ}, which enhances SAC's performance via data augmentation \citep{yarats2021image}. We replace the standard DrQ architecture of \citet{yarats2021image} with a larger one recommended for faster learning \citep{nikishin_primacy_2022, jaxrl}, which consists of an encoder followed by two MLPs for the actor and two critics. The encoder consists of four convolutional layers with output feature maps $\{ 32, 64, 128, 256 \}$ and strides $\{ 2, 2, 2, 2 \}$, respectively, followed by a linear projection to a 50-dimensional output, layer normalization \citep{ba2016layernormalization}, and then a tanh activation; the MLPs consist of two dense 256-dimensional layers, with output layers defined exactly as is done above with SAC. As in SAC, we use decoupled architectures for both the actor and the critics, unlike the original baseline, in which the encoder is shared.

%% file: 4_results.tex
\begin{figure} [t!]
    \centering
    \includegraphics[width=1\linewidth]{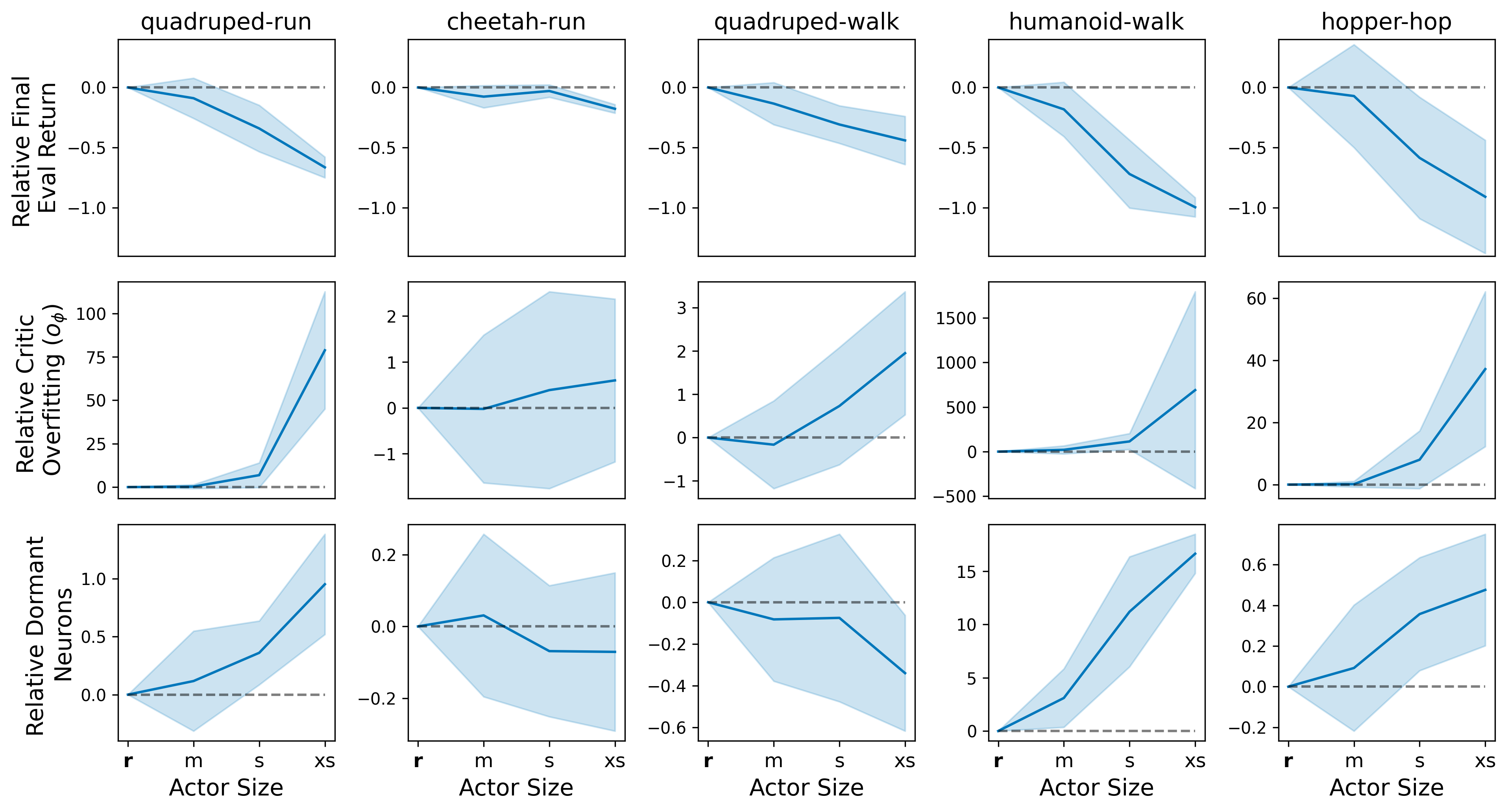}
    \caption{{\bf Decreasing the size of the actor in SAC decreases performance (top row) and increases overfitting in the critics}, as measured by $o_{\phi}$ \citep[middle row]{nauman_overestimation_2024} and dormant neurons \citep[bottom row]{sokar2023dormant}. In the top row, the y-axis is kept fixed to show the relative performance impacts across environments; this becomes impractical for the metrics in the middle and bottom row. We report the final performance, where the solid lines indicate the mean, while the shaded area represents the $95\%$ confidence interval, as computed from $10$ seeds. In all rows we report values relative to the default baseline.}
    \label{fig:smallActorImpactSAC2}
\end{figure}

\section{The impact of small actors}
\label{sec:impactOfSmallActors}
We begin by evaluating the impact on performance resulting from reduced actors. We use default hyperparameters \citep{haarnoja2018sac2} and keep the critic architecture fixed, but explore reducing the dimensionality of the actor. We denote by {\bf r} (for {\bf r}egular) the default dimensionality discussed above and use the following labels to indicate the dimensionality of the two dense hidden layers in SAC: {\bf m}: 128; {\bf s}: 32; {\bf xs}: 8. The latter correspond to network weight numbers that are $32\%$, $5\%$, and $1\%$ that of the default actor, respectively. In DrQ, we follow the same procedure as with SAC when modifying the projection MLPs, which leads to a corresponding parameter reduction as mentioned earlier in the MLPs. However, to further reduce expressivity, we also reduce the number of convolutional layers as follows: {\bf m}: $\{ 32, 64, 128 \}$; {\bf s}: $\{ 32, 64 \}$; {\bf xs}: $\{ 32 \}$. This results in more overall parameters for the DrQ actors, but this increase is at the encoder representation level, not at the direct policy level. To quantify the impact of the reduced actors, we report values relative to the baseline values. For instance, for a measure $X_s$ obtained with the {\bf s} actor, we report $\frac{X_s-X_r}{X_r}$, where $X_r$ is the value obtained with the default actor.

In the top row of \cref{fig:smallActorImpactSAC2} we evaluate the impact on performance when reducing the size of these layers and can see a clear degradation in performance across all environments. We additionally measure $o_{\phi}$ on the critics, introduced by \citet{nauman_overestimation_2024} as a measure of overfitting, defined as $o_{\phi} := \frac{\mathbb{E}_{\dataset_V}TD_{\phi}}{\mathbb{E}_{\dataset}TD_{\phi}}$. Here, $\dataset_V$ is a validation dataset of size $11,000$, containing data sampled from a training run with a regular unmodified SAC agent, trained with a different random seed, and $TD_{\phi}$ is the temporal difference error. Higher values of $o_{\phi}$ are indicative of overfitting which, as seen in the middle row of \cref{fig:smallActorImpactSAC2}, are inversely correlated with the size of the actor. Finally, we report the fraction of dormant neurons, defined as the proportion of neurons that are $0$ for every data point in the validation buffer, where higher levels of dormancy is associated with a loss of plasticity \citep{sokar2023dormant, Lyle2024DisentanglingTC, klein2024plasticitylossdeepreinforcement}. In the bottom row of \cref{fig:smallActorImpactSAC2} we see that the fraction of dormant neurons tends to be inversely correlated with actor size and performance, particularly for the environments where the performance loss is greatest, although to a lesser extent than $o_{\phi}$.

\begin{figure} [!h]
    \centering
    \includegraphics[width=1\linewidth]{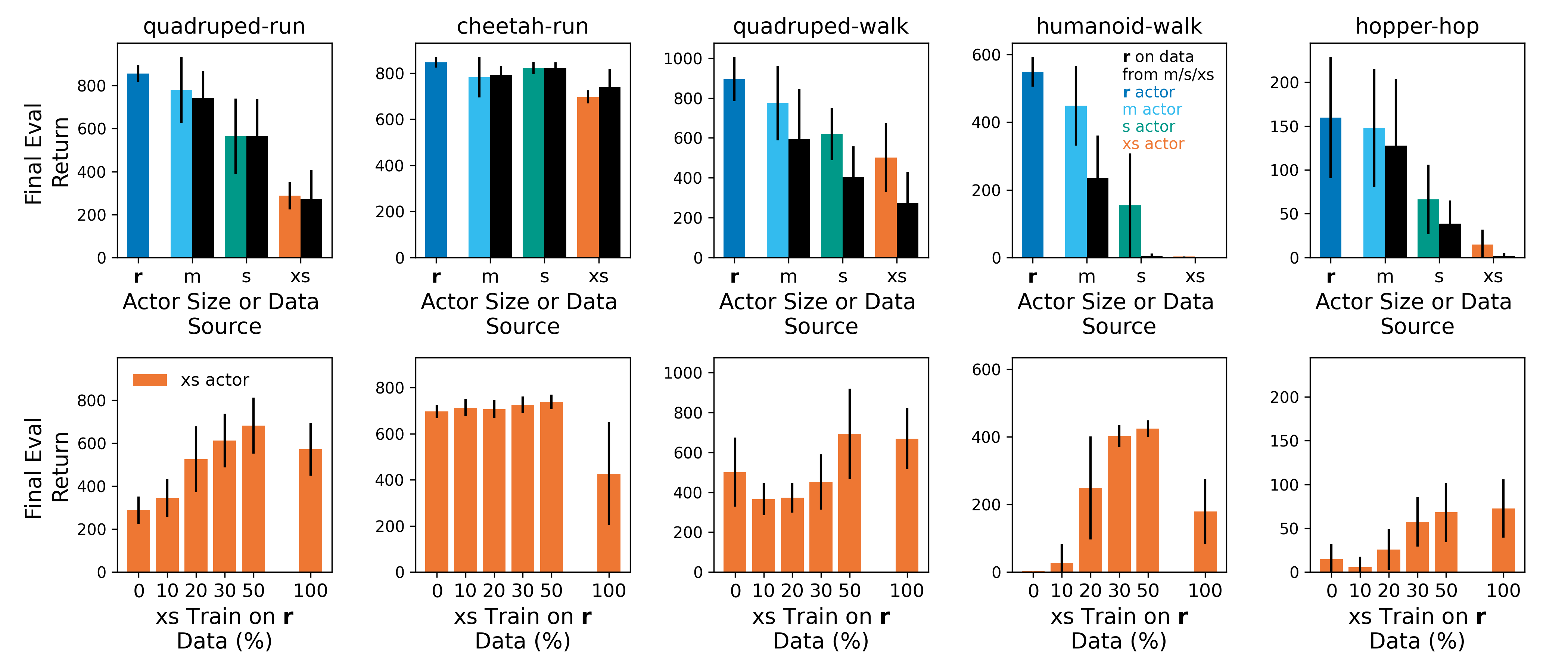}
    \caption{{\bf Evaluating the impact of data quality collecting by actors of varying sizes.} Top row: the black bars denote training regularly-sized $\pi_r$ on data collected by one of the smaller actors, while the colored bars indicate each actor trained on its own data. Bottom row: the smallest ({\bf xs}) actor trained on data from the largest ({\bf r}) actor, for varying fractions of training length. In both rows we report the final performance, where the bars indicate the mean, while the error bars represent the $95\%$ confidence intervals, as computed from $10$ independent seeds.}
    \label{fig:tandemExperiments}
\end{figure}

Figure~\ref{fig:smallActorImpactDrQ} illustrates the impact of actor reduction in DrQ, where in the pixel-based case we focus on evaluation return and critic overfitting as measured by $o_\phi$. As with SAC, we see a decrease in performance with smaller actors, as well as a general increase in $o_{\phi}$.

\subsection{Smaller actors collect worse data}

Training overparameterized neural networks on small datasets is a common cause for overfitting. Given that the actor is in charge of data collection and reducing its expressivity results in overfit critics, we continue our investigation by evaluating the quality of the data gathered by differently-sized actors. For this, we evaluate training on data collected by separate, and differently-sized, actors. Specifically, we train the regularly-sized actor $\pi_r$ with data provided by one of the smaller actors, where the data collection exactly mimics that obtained by the smaller actor during training. This is depicted by the black bars in the top row of \cref{fig:tandemExperiments}, where we can see the performance to be clearly correlated with actor size.

\citet{nikishin_primacy_2022} demonstrated the tendency of RL agents to overfit to early experience, affecting their plasticity and downstream performance.
It is thus worth considering whether the quality of the training data on an actor is most important in the early stages of training. To evaluate this, in the bottom row of \cref{fig:tandemExperiments} we explore training the smallest actor ($\pi_{xs}$) on data provided by the $\pi_r$ actor, again matching the data collection of the smaller actor. Our analyses here explore using the data from $\pi_r$ for only a fraction of training, and then switching to data collected by $\pi_{xs}$ itself. 
As more data is collected from the bigger actor $\pi_r$, the performance of $\pi_{xs}$ generally improves. We note that using all of the data from $\pi_r$ (i.e. at 100\%) sometimes results in degraded performance; we hypothesize that this may be due to the tandem effect observed by \citet{ostrovski2021difficulty}. Overall, we see an improvement in performance in the environments most impacted by reducing the size of the actor for SAC. With DrQ, we do not see a pronounced effect when training the smallest actor on data from a regular-sized actor (see \cref{fig:smallActorDataImpactDrQ}), but similarly, a small trend may be observed for environments with the biggest degradation in performance with reduced actor sizes.

\begin{figure} [!h]
    \centering
    \includegraphics[width=1\linewidth]{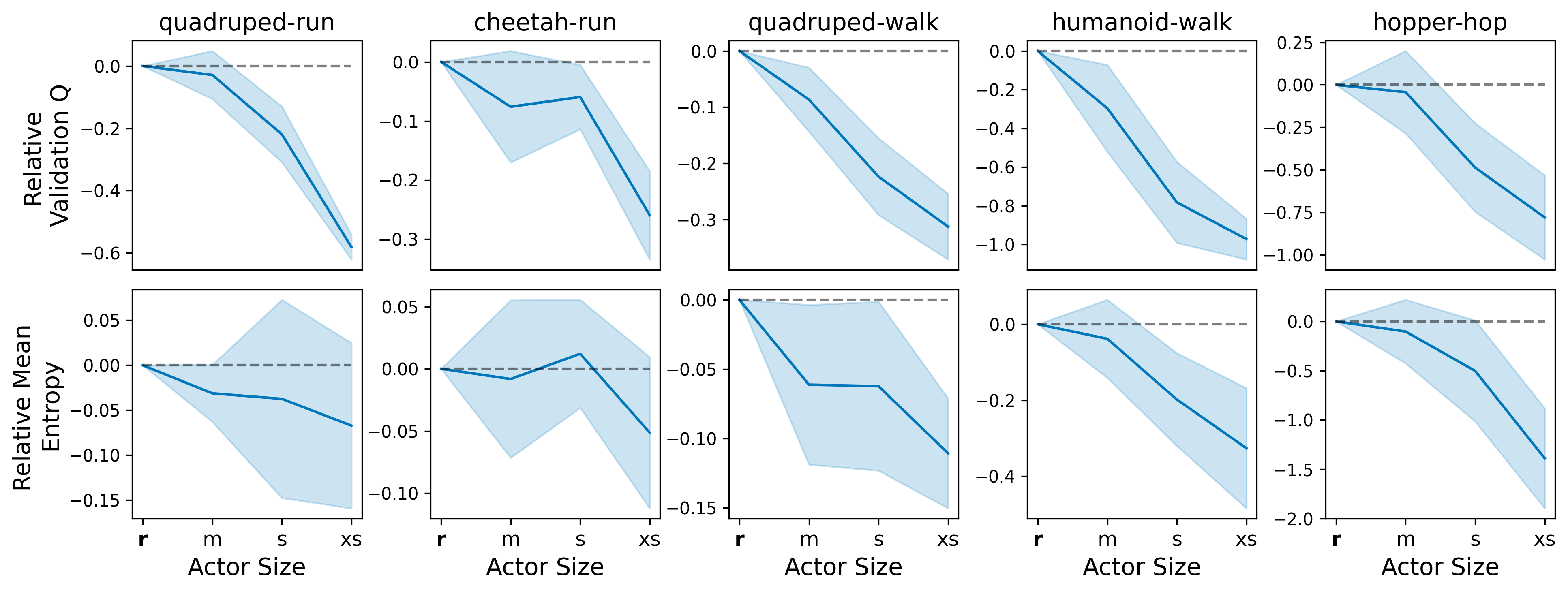}
    \caption{{\bf Decreasing the size of the actor results in $Q$-value underestimation and reduced policy entropy.} In the top row we estimate the average $Q$-values on a batch of data gathered during evaluation, and plot the values relative to the baseline {\bf r}. In the bottom row we compute the entropy of the policy $\pi$ and plot the values relative to the entropy of the baseline {\bf r}. In both cases we report the values obtained at the end of training, where the solid line represents the mean with shaded areas indicating 95\% confidence intervals, computed over 10 independent seeds.}
    \label{fig:QValueUnderestimation}
\end{figure}

\subsection{Smaller actors result in critic underestimation}
The reduction in the quality of data gathered by small actors can possibly be attributed to under-exploration of the state space. This can often be a consequence of an overly-conservative critic which under-estimates values, as well as a low-entropy actor with low diversity in action selection. In \cref{fig:QValueUnderestimation} we compare the average critic validation $Q$-values (computed on the  same validation dataset) as well as the entropy of the actor's action distribution $\pi$ of the smaller actors relative to the regularly-sized actor, and confirm that smaller actors result in $Q$-value underestimation, as well as a reduction in entropy during training (see \cref{fig:QValueUnderestimationTrainingCurves} for the same comparison throughout training). The observed underestimation is interesting, given that it stands in contrast to the {\em over-estimation} that's commonly addressed in many popular algorithms \citep{hasselt2010double,hasselt2016deep,fujimoto2018addressing}.

%% file: 4.5_fixes.tex
\begin{figure} [!t]
    \centering
    \includegraphics[width=1\linewidth]{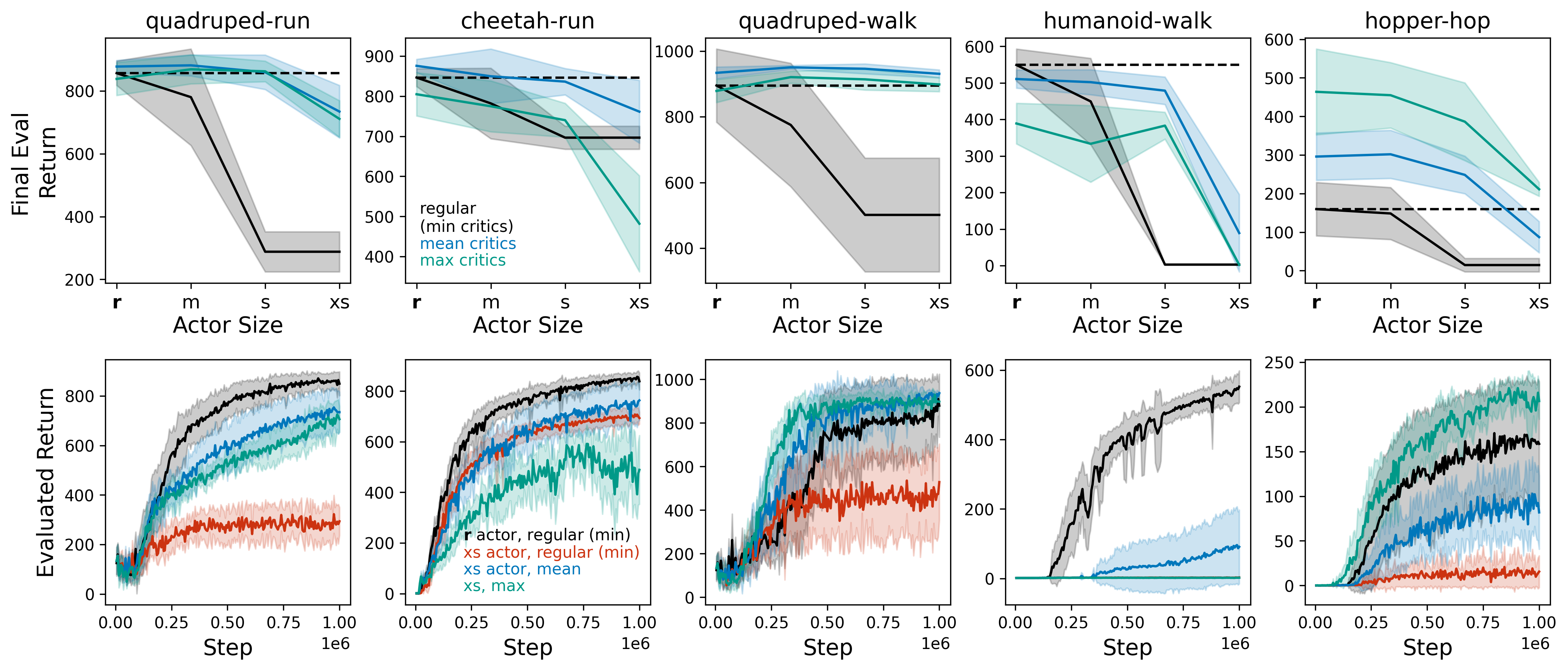}
    \includegraphics[width=0.9\linewidth]{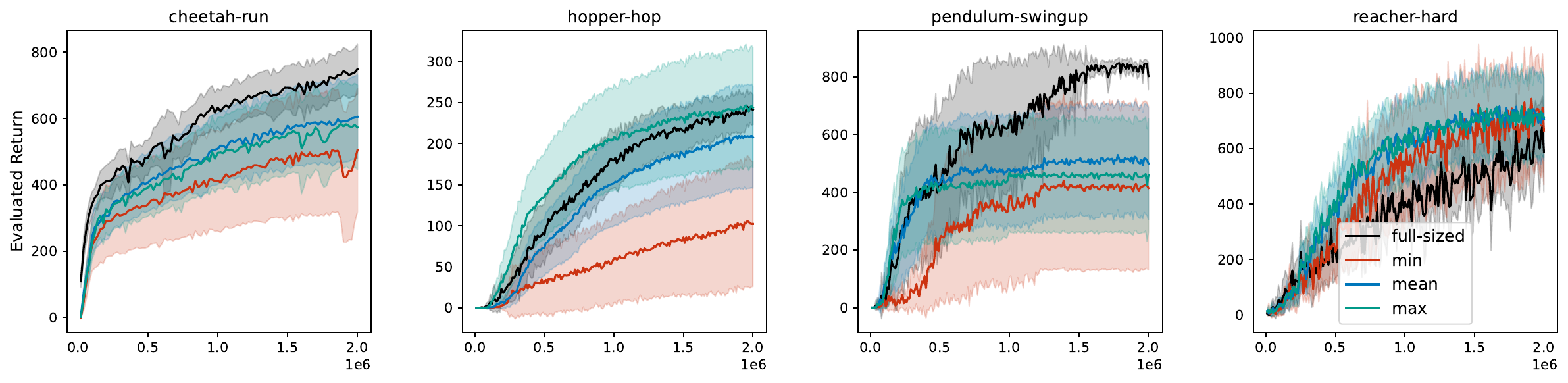}
    \caption{{\bf Taking the mean or max of the two critics can empower smaller actors} in SAC (top and middle rows) and on the smallest actor in DrQ (bottom row). Replacing the minimums in equations \ref{eqn:sac_jq_min} and \ref{eqn:sac_jpi_min} with mean and max can help reduce $Q$-value underestimation and boost performance. The top row displays final performance while the bottom two rows display performance throughout training, with solid lines indicating average over 10 seeds, and shaded areas 95\% confidence intervals. In DrQ, the performance over the non-full-sized settings is computed using 20 seeds to account for higher observed variance.}
    \label{fig:maxMeanCritics}
\end{figure}

\section{Empowering small actors}
\label{sec:empoweringSmallActors}

The results from the last section suggest that the  performance reduction resulting from the use of small actors  is largely due to poor data collection, which in turn appears to be a consequence of value underestimation and low action variability. In this section we explore a variety of approaches for strengthening small actors.

\subsection{Average and maximal critics}
\label{sect:avgMaxCritics}
We begin by a simple modification to the original SAC losses to directly address the observed value underestimation. Specifically, we replace the minimization of the two independent $Q$ estimates in equations \ref{eqn:sac_jq_min} and \ref{eqn:sac_jpi_min} with either their mean ($\textrm{avg}(Q_{\phi_1}, Q_{\phi_2})$) or their maximum ($\max (Q_{\phi_1}, Q_{\phi_2}))$.  As can be seen in the top and middle rows of \cref{fig:maxMeanCritics} and the top row of \cref{fig:relqs}, this approach can be quite effective at boosting the performance of small actors in SAC, sometimes even improving over the minimization approach with the regular sized model (e.g. hopper-hop). The bottom row of \cref{fig:relqs} confirms that this technique does increase the validation value estimates. As can be seen in \cref{fig:plasticityMetricsCritic}, we find that the mean and the max approaches also improve several overfitting and plasticity metrics in the critics, most notably $o_{\phi}$ and the rank of the last hidden layer \citep{kumar2021implicit, nauman_overestimation_2024}. However, they do not appear to have a notable impact on these metrics in the actor (see \cref{fig:plasticityMetricsActor}). The results on the smallest actor on DrQ (bottom row of \cref{fig:maxMeanCritics}) display a similar performance trend, although the results are less pronounced. We also observe a corresponding trend with an increase in validation Q values with the mean and max approaches in DrQ in \cref{fig:smallActorResetMeanMax}.

\subsection{Critic regularization}
\begin{figure} [!t]
    \centering
    \includegraphics[width=1\linewidth]{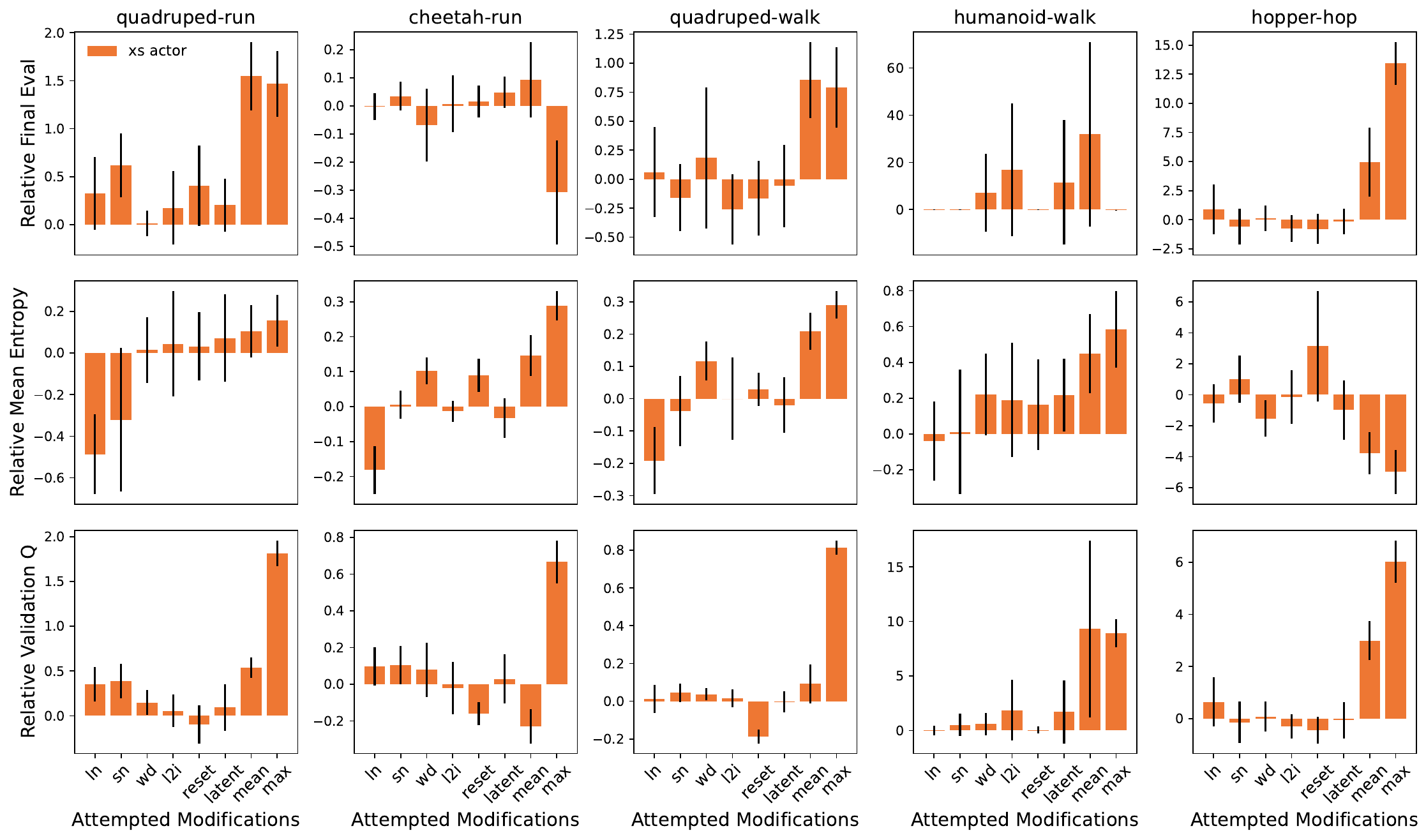}
    \caption{{\bf Impact of attempted modifications on the final performance (top), mean entropy of the actor's action distribution (middle), and validation $Q$-value estimation (bottom) of the smallest actor in SAC}. The values are relative to the smallest actor in unmodified SAC. Bars indicate mean with error bars denoting 95\% confidence intervals, computed over 10 independent seeds.}
    \label{fig:relqs}
\end{figure}

Prior work has shown that resetting or regularizing the critic in particular improves plasticity \citep{ma_revisiting_2023, nikishin_primacy_2022, liu2021regularization} and can help mitigate value-estimation issues, particularly in the case of layer normalization, \citep{nauman_overestimation_2024}, albeit with overestimation. Given both the increased overfitting observed in the critics, and how much value estimation is affected by smaller actors (see \cref{fig:QValueUnderestimation}), we investigate whether critic regularization alone can be effective mitigating this impact by applying a number of regularization techniques, focusing on SAC: 
(a) {\bf Layer Normalization} \citep{ba2016layernormalization}; (b) {\bf Spectral Normalization} \citep{miyato2018spectral}; (c) {\bf weight decay} \citep{laarhoven2017regularization} with a regularization value of $0.01$ \citep{li2023efficientdeepreinforcementlearning}; (d) {\bf L2 distance from initialization} \citep{kumar2024maintaining}: with a value of $1 \times 10^{-7}$ after tuning on the range $[5 \times 10^{-8}, 1 \times 10^{-4}]$ in increments of $0.5$  with quadruped-run; and (e) {\bf Network resets:} resetting neural network layers during training \citep{nikishin_primacy_2022}. We apply layer normalization and spectral normalization to the second hidden layer in the critics, and we reset only the output layer of the critics every $50$K steps. Although many of these methods do appear to help with mitigating value under-estimation (bottom row of \cref{fig:relqs}), they do not appear to help much with performance (top row of \cref{fig:relqs} and \cref{fig:RegularizationEffects}). For DrQ, we investigate resetting the MLP of the critics \citep{nikishin_primacy_2022} for the smallest actors, and similarly do not see a notable rescue effect (see \cref{fig:smallActorResetMeanMax}).

\subsection{Addressing bias in the critic via actor representations}

In asymmetric actor-critics methods, imbalances in information received by the critic versus the actor can lead to biased gradients that may negatively impact performance; \citet{baisero_unbiased_2022} and \citet{Lyu2022ADU} propose to alleviate this by giving the (limited) information received by the actor as additional input into the critic. In our case, the critics do not receive privileged information over the actor, but we theorize that a similar effect may be occurring within the policy network due to potentially impacted information flow through the smaller actors. We attempt a similar bias correction by concatenating the latent state of the final hidden layer of the actor as input to the final hidden layer of the critics. The latent state is first projected through an untrained neural network layer to a size of $8$ to maintain consistency across actor sizes. As shown in the top row of \cref{fig:relqs} and the bottom row of \cref{fig:RegularizationEffects}, the bias correction performs similarly to other attempted critic regularization methods in SAC.

%% file: 5_conclusion.tex
\section{Discussion}
\label{sec:conclusion}
Real-world problems are often subject to constraints such as latency, model size, and interpretability, which are largely absent in the academic benchmarks where machine learning solutions are developed.
As such, it is imperative that we develop the necessary techniques for training reinforcement learning agents under such limitations. The use of small actors can help reduce latency, memory, and inference costs, and can help improve interpretability; these are all practical considerations, as ultimately it is a trained actor which will be deployed for action selection. Our work demonstrates that na{\" i}vely shrinking the actor can result in value underestimation, poor data collection, and ultimately degraded performance. We evaluated a number of approaches for mitigating this deterioration and found the most effective to be simply replacing the min operation with a mean or max when combining the values of the two critics (\cref{sect:avgMaxCritics}).

\begin{figure} [!t]
    \centering
    \includegraphics[width=1\linewidth]{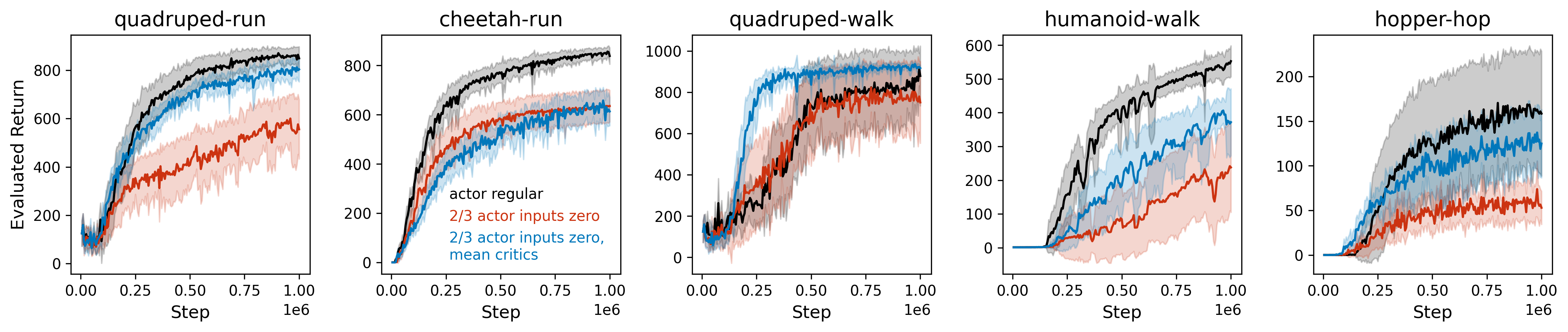}
    \caption{{\bf Taking the mean of the two critics can help deal with partial observability in the actor.} We zero-out 2/3 of the inputs into the actor and compare the performance when using the min or mean of the two critics.}
    \label{fig:partialObservability}
\end{figure}

It is often necessary to provide the actor with less information than the critic, as was employed by \citet{vasco_super-human_2024} to better match the inputs used by humans. In \cref{fig:partialObservability} we explore whether this additional type of limitation on the actor may have a similar effect to what we observe when decreasing the size of the actor. To test this, we zero out two-thirds of the actor inputs in SAC (retaining every third dimension) and find that taking the mean of the two critics - rather than the minimum - alleviates performance loss here as well. Of note, the alleviation is more pronounced in the same environments where underestimation mitigation helped the most with smaller actors (\cref{fig:maxMeanCritics}). This suggests that addressing underestimation in SAC can additionally help mitigate the challenges arising from partial observability.

Figure~\ref{fig:plasticityMetricsActor} suggest that smaller actors result in larger parameters and reduced effective rank, which are often tied to optimization difficulties; figure~\ref{fig:plasticityMetricsCritic} suggests that these effects are less pronounced on the critics. Interestingly, the most effective technique we found for mitigating value underestimation seems to have little impact on the actor's parameter norms and effective rank, but does seem to play an important role on the critics.

In general, developing a greater understanding between optimization, exploration, expressivity, and estimation accuracy will lead to more robust and reliable reinforcement learning agents. While our work has focused on the case of small actors, the insights provided help strengthen our collective understanding of these learning dynamics. Addressing overestimation in AC methods by taking the minimum of estimated Q values has been a continuing trend - for example, with Deep Deterministic Policy Gradient  \citep[{DDPG};][]{lillicrap2015continuous} being followed by Twin-Delayed DDPG \citep[{TD3};][]{fujimoto2018addressing}. However, our work contributes to findings showing that this approach contributes to underestimation, which warrants further consideration particularly in settings where data collection is more challenging \citep{NEURIPS2019_a34bacf8, 10.3389/fnbot.2022.1081242, He_2020}. Further, all these considerations are aligned with the continued relevance of the exploration-exploitation dilemma \citep{10.3389/fnbot.2022.1081242, Sutton1998}, which has been explored via Thompson sampling-like techniques \citep{ishfaq2025langevinsac} and through information gain maximization \citep{sukhija2025maxinforl}.

\paragraph{Limitations} Our empirical investigations were mostly focused on SAC evaluated on DMC with feature-based observations. Although we did conduct subsets of our analyses on DrQ with the more challenging pixel-based observations, further evaluations on different benchmarks and agents would be necessary to strengthen the generality of our claims. For consistency and computational considerations, in our work we used the default hyper-parameters of the baseline models for all experiments; however, RL agents can often be sensitive to hyper-parameter choices \citep{ceron2024consistency}, so ideally one would perform a hyper-parameter search for each the various settings considered, although this can be computationally prohibitive.

\subsubsection*{Broader impact statement}
\label{sec:broaderImpact}

This paper presents work whose goal is to advance the field
of Reinforcement Learning. There are many potential societal
consequences of our work, none which we feel must be
specifically highlighted here.

\subsubsection*{Acknowledgements}
We thank Johan Obando-Ceron, Roger Creus Castanyer, Gandharv Patil, Ayoub Echchahed, Ali Saheb Pasand, Joao Madeira Araujo, and the RLC technical and senior reviewers for providing helpful feedback on our work.

We would also like to thank the Python community \citep{van1995python, 4160250} for developing tools that enabled this work, including NumPy \citep{harris2020array}, Matplotlib \citep{hunter2007matplotlib}, Jupyter \citep{2016ppap}, and Pandas \citep{McKinney2013Python}.

%% file: 6_appendix.tex
\beginSupplementaryMaterials
\section{Extra results}

We include extra results that support the claims made in the main sections, but are not necessary to properly follow the paper.

\begin{figure} [!h]
    \centering
    \includegraphics[width=\linewidth]{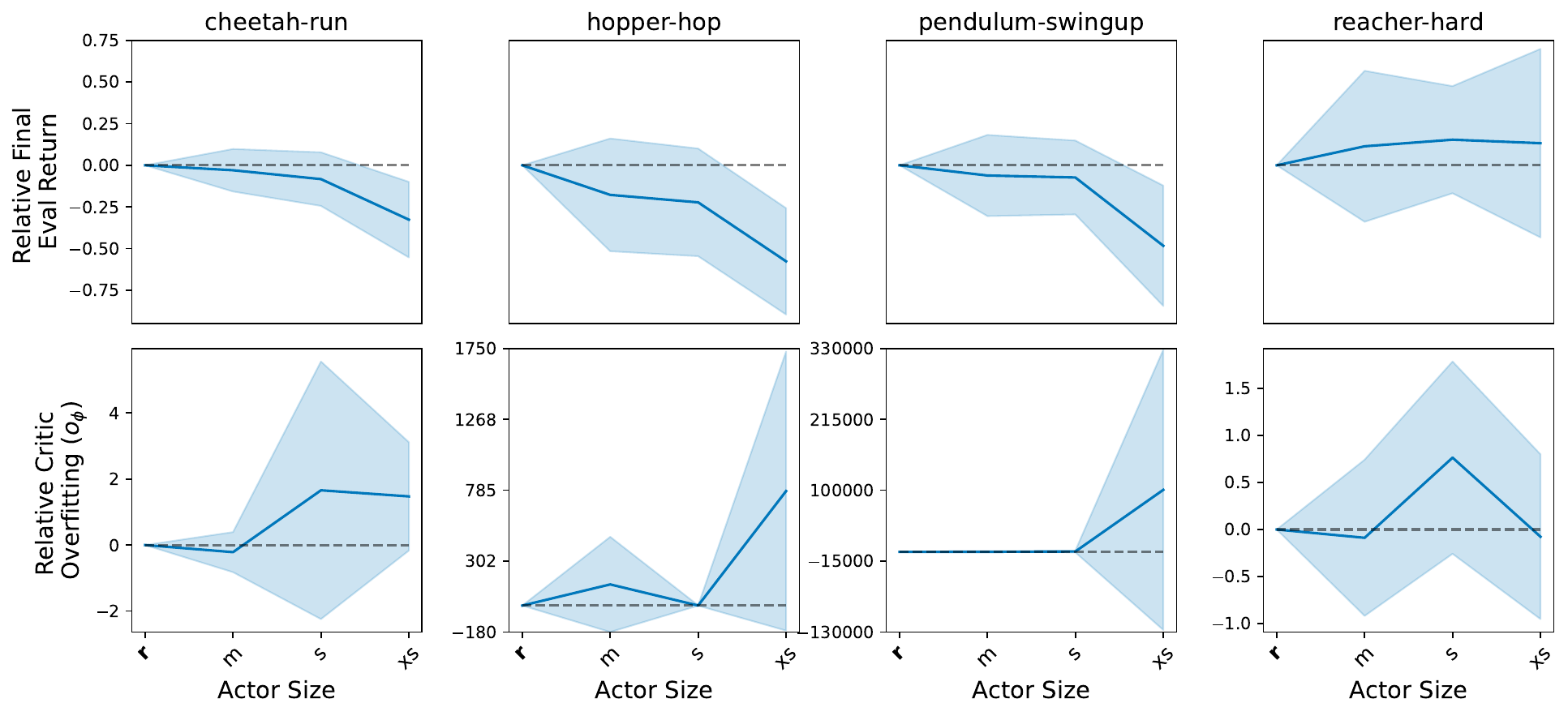}
    \caption{{\bf Decreasing the size of the actor in DrQ decreases performance (top row) and increases overfitting in the critic}, as measured by $o_{\phi}$ \citep[bottom row]{nauman_overestimation_2024}. The solid lines represent mean performance, while the shaded area represents the $95\%$ confidence interval, computed across $10$ seeds. In all rows we report values relative to the default baseline.}
    \label{fig:smallActorImpactDrQ}
\end{figure}

\begin{figure} [!h]
    \centering
    \includegraphics[width=\linewidth]{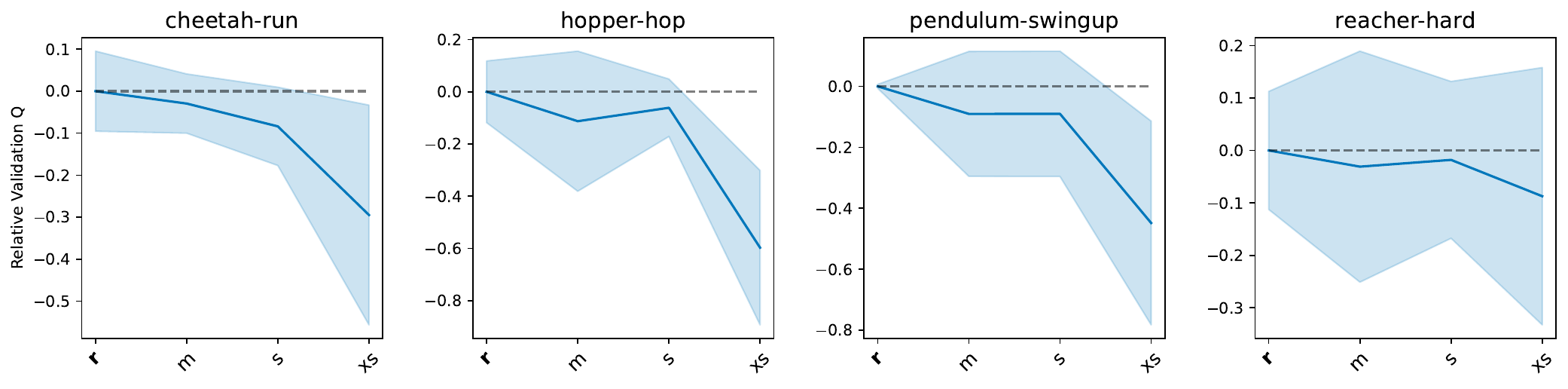}
    \caption{{\bf Decreasing the size of the actor in DrQ decreases validation Q values}, as described in \cref{fig:QValueUnderestimation}. We report Q values at the end of training relative to the default baseline. The solid lines represent mean performance, while the shaded area represents the $95\%$ confidence interval, computed across $10$ seeds.}
    \label{fig:smallActorImpactDrQ}
\end{figure}

\begin{figure} [!h]
    \centering
    \includegraphics[width=\linewidth]{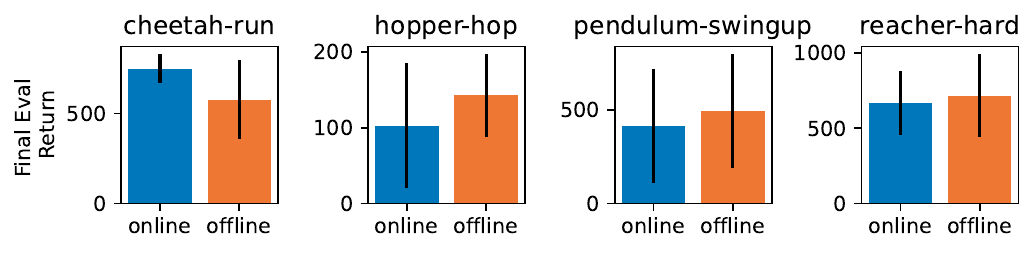}
    \caption{{\bf Training the smallest actor on data collected by the largest, high-performing actor does not appear to lead to a clear improvement in performance in DrQ across a suite of environments,} possibly due to high variance, although an improvement trend might be suggested for hopper-hop and pendulum-swingup. The blue bars are the default final performances of the smallest actors, and the orange bars are the final performances of the smallest actors trained on data collected by a regular-sized actor. Results are aggregated across 10 seeds, and the error bars are the confidence $95\%$ confidence intervals.}
    \label{fig:smallActorDataImpactDrQ}
\end{figure}

\begin{figure} [!h]
    \centering
    \includegraphics[width=\linewidth]{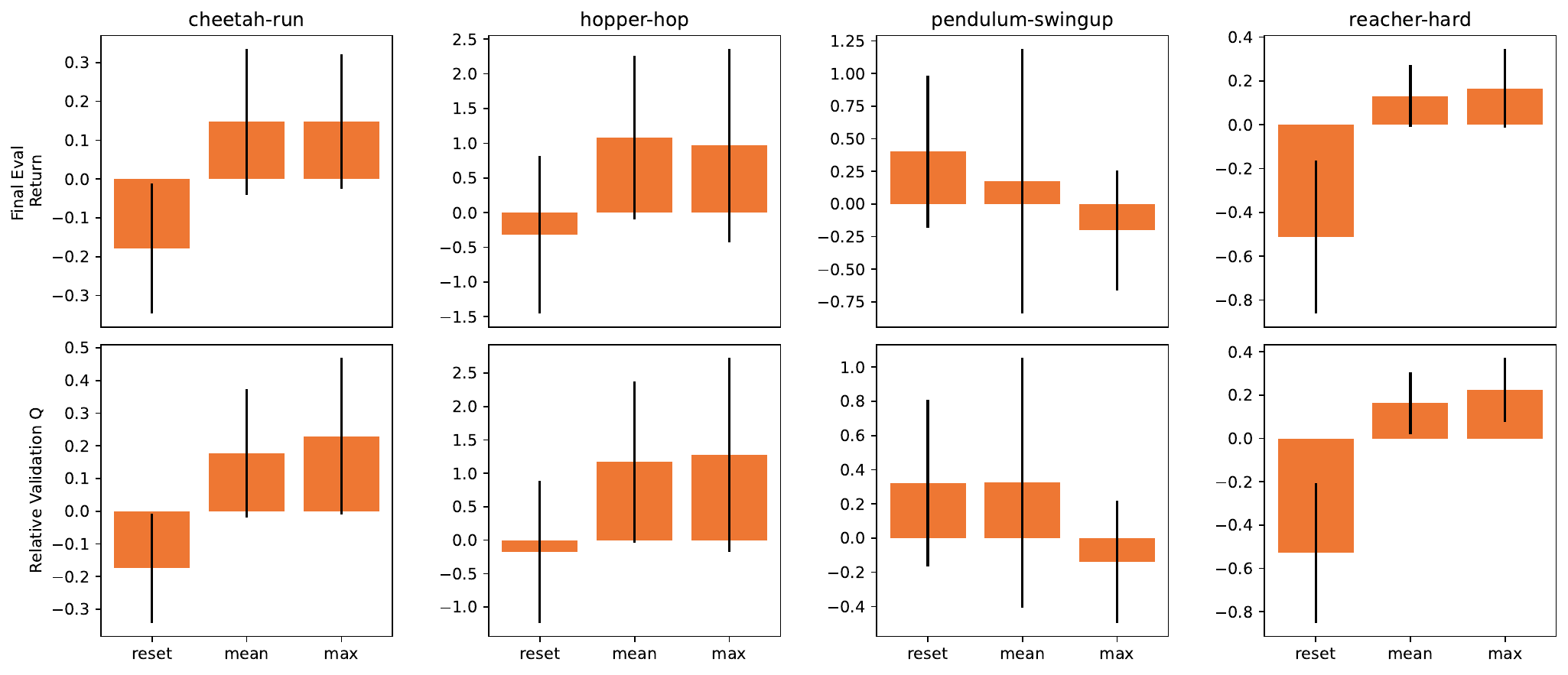}
    \caption{{\bf Impact of various modifications on final performance and validation $Q$-values in DrQ.} In the resetting experiments, e reset every $100,000$ steps, and only reset the critics. As in \citep{nikishin_primacy_2022}, we only reset the MLP of the critics, and leave the encoder untouched. The error bars indicate the $95\%$ confidence interval, computed over $10$ seeds.}
    \label{fig:smallActorResetMeanMax}
\end{figure}

\begin{figure} [!h]
    \centering
    \includegraphics[width=1\linewidth]{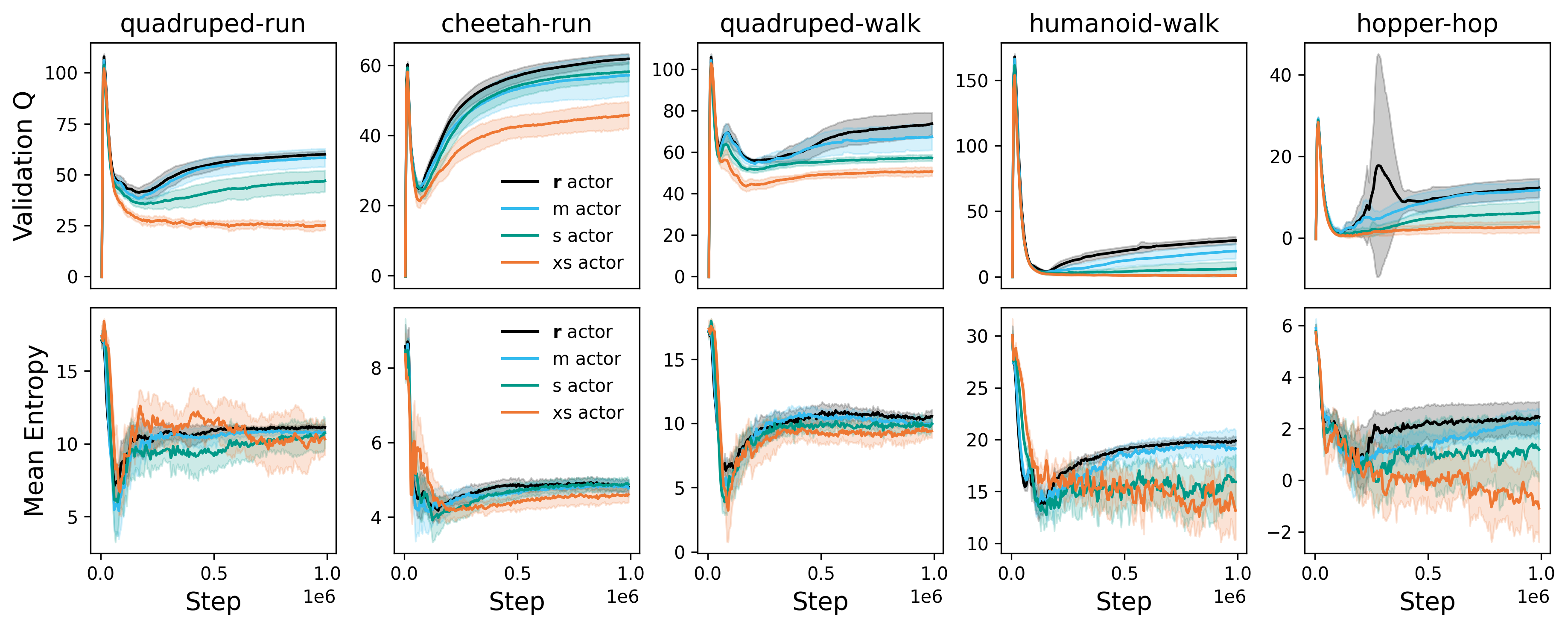}
    \caption{{\bf Decreasing the size of the actor results in $Q$-value underestimation and reduced policy entropy.} In the top row we estimate the average $Q$-values on a batch of data gathered during evaluation, and plot the values relative to the baseline {\bf r}. In the bottom row we compute the entropy of the policy $\pi$ and plot the values relative to the entropy of the baseline {\bf r}. In both cases the solid line represents the mean with shaded areas indicating 95\% confidence intervals, computed over 10 independent seeds.}
    \label{fig:QValueUnderestimationTrainingCurves}
\end{figure}

\begin{figure} [h!]
    \centering
    \includegraphics[width=1\linewidth]{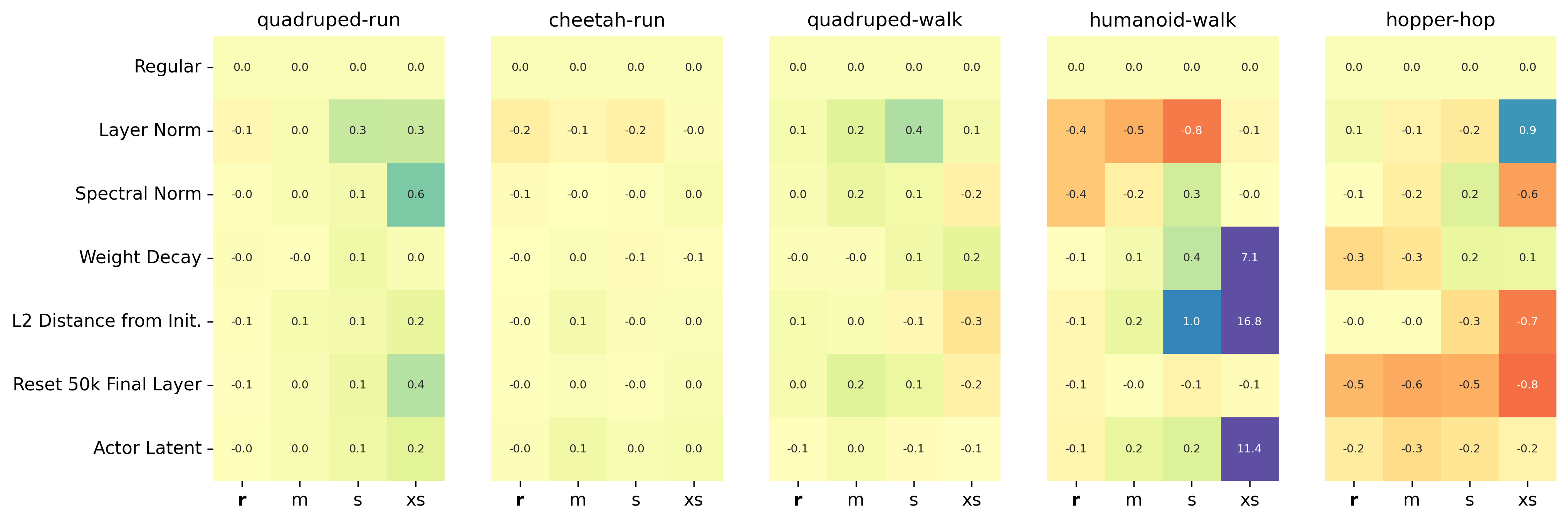}
    \caption{{\bf Impact of critic regularizations on downstream performance with actors of varying sizes.} Each table row corresponds to one of the normalization mechanisms explored, each column indicates the actor size used, and the value in each cell denotes the change relative to the unnormalized version (top row). In most environments there is little change, although in humanoid-walk some regularization techniques do appear to mitigate the performance loss from smaller actors.}
    \label{fig:RegularizationEffects}
\end{figure}

\begin{figure} [ht]
    \centering
    \includegraphics[width=1\linewidth]{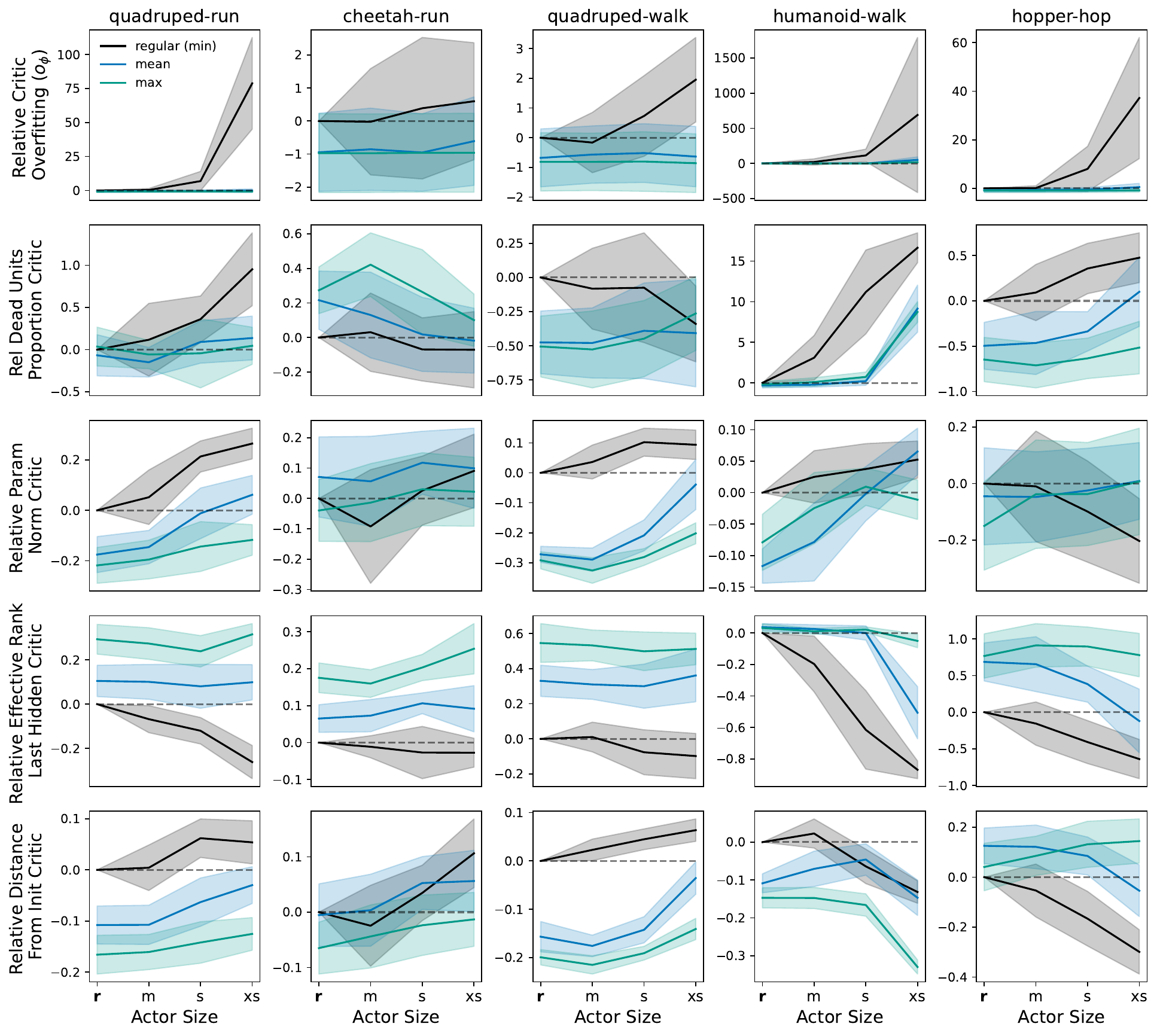}
    \caption{{\bf The impact of small actors on a number of metrics related to plasticity as measured on the critics.} We also evaluate these metrics when using the mean and max of the two critics, as discussed in \cref{sect:avgMaxCritics}.}
    \label{fig:plasticityMetricsCritic}
\end{figure}

\begin{figure} [ht]
    \centering
    \includegraphics[width=1\linewidth]{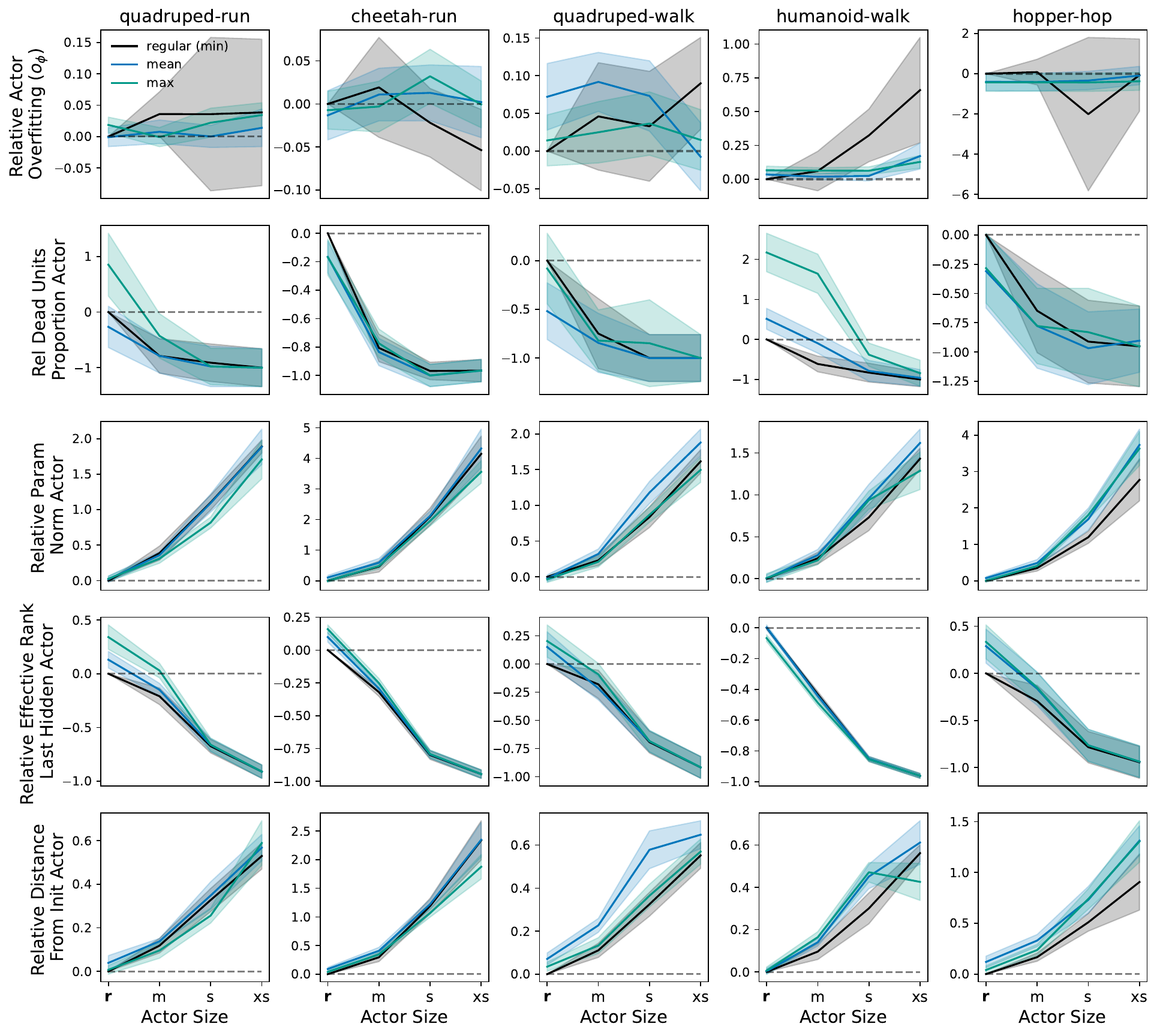}
    \caption{{\bf The impact of small actors on a number of metrics related to plasticity as measured on the actor.} We also evaluate these metrics when using the mean and max of the two critics, as discussed in \cref{sect:avgMaxCritics}. We define $o_{\phi}$ on the actor as $o_{\phi} := \frac{\mathbb{E}_{\dataset}H}{\mathbb{E}_{\dataset_V}H}$, where $H$ is the entropy of the actor's action distribution, and $\dataset_V$ is a validation dataset.}
    \label{fig:plasticityMetricsActor}
\end{figure}